\documentclass[sn-mathphys,Numbered]{sn-jnl}

\usepackage{graphicx}%
\usepackage{multirow}%
\usepackage{amsmath,amssymb,amsfonts}%
\usepackage{amsthm}%
\usepackage{mathrsfs}%
\usepackage[title]{appendix}%
\usepackage{textcomp}%
\usepackage{manyfoot}%
\usepackage{booktabs}%
\usepackage{algorithm}%
\usepackage{algorithmicx}%
\usepackage{algpseudocode}%
\usepackage{listings}%
\usepackage{pslatex}
\usepackage[bottom]{footmisc}
\usepackage{xfrac}
\usepackage{multirow}
\usepackage{mathtools}
\usepackage{float}
\usepackage{bm}
\usepackage{colortbl}
\usepackage[table,xcdraw]{xcolor}

\begin{document}

\title[Article Title]{Multi-Class Anomaly Detection based on Regularized Discriminative Coupled hypersphere-based Feature Adaptation}


\author*[1]{\fnm{Mehdi} \sur{Rafiei}}\email{rafiei@ece.au.dk}

\author[1]{\fnm{Alexandros} \sur{Iosifidis}}\email{ai@ece.au.dk}

\affil*[1]{\orgdiv{DIGIT, Department of Electrical and Computer Engineering}, \orgname{Aarhus University},\city{Aarhus}, \country{Denmark}}


\abstract{In anomaly detection, identification of anomalies across diverse product categories is a complex task. This paper introduces a new model by including class discriminative properties obtained by a modified Regularized Discriminative Variational Auto-Encoder (RD-VAE) in the feature extraction process of Coupled-hypersphere-based Feature Adaptation (CFA). By doing so, the proposed Regularized Discriminative Coupled-hypersphere-based Feature Adaptation (RD-CFA), forms a solution for multi-class anomaly detection. By using the discriminative power of RD-VAE to capture intricate class distributions, combined with CFA's robust anomaly detection capability, the proposed method excels in discerning anomalies across various classes. Extensive evaluations on multi-class anomaly detection and localization using the MVTec AD and BeanTech AD datasets showcase the effectiveness of RD-CFA compared to eight leading contemporary methods.}




\maketitle

\section{Introduction}
\label{sec:introduction}

\begin{figure}
\centering
\includegraphics[width=0.8\columnwidth]{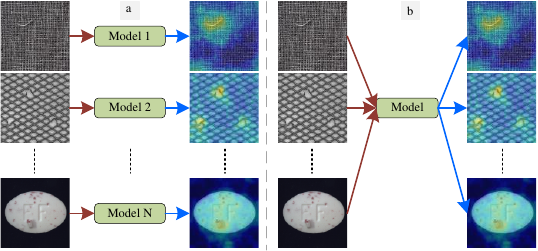}
\caption{a) Single-class anomaly detection, b) Multi-class anomaly detection.}
\label{MC_task}
\end{figure}

The increasing prevalence of anomaly detection across diverse sectors, including product inspection, medical diagnostics, and security applications, underscores its important role in many applications \cite{lee2022cfa}. Multi-class anomaly detection is one of the most intricate tasks within this domain. Traditionally, tackling anomaly detection involves creating effective models that can accurately represent the distribution of normal samples, thereby classifying any deviations from this distribution as anomalies. Many existing methods attempt to address the multi-class anomaly detection task by employing separate models for distinct object classes \cite{zavrtanik2021draem}, \cite{defard2021padim}, \cite{lee2022cfa}, \cite{gudovskiy2022cflow}, and \cite{deng2022anomaly}. However, as the number of classes increases, this one-class-one-model strategy (depicted in Figure \ref{MC_task}-a) becomes resource-intensive and can significantly strain computational capabilities.

A viable solution to this challenge lies in employing a single model for anomaly detection across diverse object classes (Figure \ref{MC_task}-b). In this approach, the training data encompasses normal samples from various categories, and the developed model is tasked with anomaly detection across all these categories without the need for fine-tuning. We believe that accurate discrimination among distinct classes is central in multi-class anomaly detection scenarios, as the methods lacking such discrimination ability tend to generate false positive predictions \cite{you2022unified,kirchheim2022multi}, indicating their inability to focus on the specific normal features that differentiate each class from the rest. Addressing this discrimination deficit is critical for enhancing the reliability of anomaly detection methodologies in complex real-world settings. Furthermore, given the diverse range of normal features present in a multi-class dataset, there exists a possibility that methods based on reconstruction neural networks could accurately reconstruct both normal and anomalous samples \cite{you2022unified}. This situation can hinder the detection of anomalies as the model tries to distinguish between the two.

The multi-class anomaly detection task is seen in \cite{tian2021self} and \cite{kirchheim2022multi} as an image-level classification task to normal and abnormal classes without anomaly localization. The method proposed in \cite{you2022unified} approaches the task as both classification and anomaly localization. To overcome the challenges of both tasks, they introduce a query decoder organized in layers, enhancing the ability of the method to model the multi-class distribution effectively. Additionally, they utilize a neighbor-masked attention module to prevent information leakage from the input to the reconstructed output. Lastly, they introduce a feature jittering strategy, compelling the model to recover accurately even when exposed to noisy inputs. 

In this paper, we propose an approach to address the challenges of multi-class anomaly detection across various products by leveraging the class discrimination ability of Regularized Discriminative Variational Auto-Encoder (RD-VAE) \cite{passalis2020variance}, added to the feature extraction process of Coupled-hypersphere-based Feature Adaptation (CFA) \cite{lee2022cfa} which was originally proposed for anomaly detection problems where normal samples come from one class. By applying RD-VAE to the patch features obtained through CFA, the method effectively captures the diverse distributions of all classes. This process enables CFA to discriminate between different classes during the anomaly detection and localization tasks, making it particularly adept at handling the challenges in multi-class anomaly detection. The method only utilizes class-label information during the training phase to establish these distributions, and the process operates independently of class labels during inference, ensuring flexibility and practical applicability.

The proposed Regularized Discriminative Coupled-hypersphere-based Feature Adaptation (RD-CFA) method is extensively evaluated against eight leading contemporary anomaly detection methods using two well-established publicly available datasets, i.e., MVTec AD \cite{bergmann2019mvtec} and BeanTech AD \cite{mishra2021vt}. Experiments show that RD-CFA improves anomaly detection accuracy and the precision of anomaly localization. 

\section{Related work}
In the proposed method, we include class discrimination properties to the CFA's feature extraction process. Such properties can be obtained by using data parametric transformations modeled by VAEs, as it was recently proposed by RD-VAE \cite{passalis2020variance}. Therefore, we modified the RD-VAE method to better discriminate between all classes in multi-class anomaly detection. This modification is included in the feature extraction process of the CFA method, enhancing anomaly detection and localization capabilities in datasets with multiple classes. Therefore, this section offers succinct overviews of the original CFA and RD-VAE methods to facilitate a comprehensive understanding.

\subsection{Coupled-hypersphere Feature Adaptation (CFA)}
CFA exploits the ability of Convolutional Neural Networks pre-trained on large datasets to extract informative features, while it tries to combat biases related to differences between the distribution of the data the network was pre-trained on and the normal samples within the target dataset by employing Transfer Learning. This ensures the concentration of image patch features around memorized features, effectively addressing the tendency to overestimate abnormality in pre-trained CNNs. 

\begin{figure}
\centering
\includegraphics[width=0.9\columnwidth]{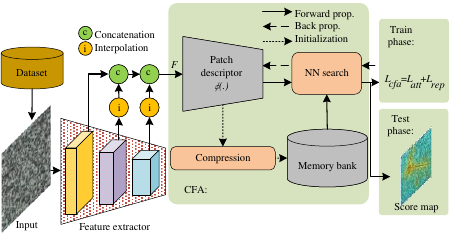}
\caption{Overall structure of the CFA model.}
\label{CFA}
\end{figure}

As depicted in Figure \ref{CFA}, patch features denoted as $F\in\mathbb{R}^{D\times H\times W}$ are obtained by inferring samples from the target dataset using a pre-trained CNN, the parameters of which are not updated. Due to varying spatial resolutions in feature maps at different CNN depths, these feature maps are interpolated and concatenated. Here, $H$ and $W$ represent the height and width of the largest feature map, while $D$ indicates the total number of dimensions of the sampled feature maps. To transform these patch features into target-oriented features, the CFA model employs an auxiliary network known as the patch descriptor network $\phi(\cdot): \mathbb{R}^D \rightarrow \mathbb{R}^{D'}$.

CFA uses a memory bank $C$ during training to store initial target-oriented features obtained exclusively from a training set containing normal samples. These features are stored based on a specific modeling procedure. The central idea behind CFA involves contrastive supervision using coupled hyperspheres created with memorized features $\textbf{c}_{l\in\{1,2,...,T\}}\in C$, where $T=H\times W$, representing the number of patch features from a single sample, as centers. This is achieved by optimizing the parameters of the patch descriptor network $\phi(\cdot)$ through the Coupled-hypersphere-based Feature Adaptation loss function, which is formed by two terms, namely the feature attractive loss $L_{f\_att}$ and the feature repulsive loss $L_{f\_rep}$. 
For the patch feature $\textbf{p}_t$, the $k$-th nearest neighbor, i.e., $\textbf{c}_t^k$, is searched through the NN search of $\phi(\textbf{p}_t)$ in $C$. Then, CFA updates the parameters of $\phi(\cdot)$ to embed $\textbf{p}_t$ close to $\textbf{c}_t^k$. To do that, $L_{f\_att}$ penalizes distances between $\phi(\textbf{p}_t)$ and $\textbf{c}_t^k$ greater than $r$, i.e.,:
\begin{equation}
    L_{f\_att} = \frac{1}{TK}\sum_{t=1}^{T}\sum_{k=1}^{K}\max\Bigg(0,D\left(\phi(\textbf{p}_t),\textbf{c}_t^k\right)-r^2\Bigg),
\end{equation}
where $K$ represents the number of nearest neighbors matching with $\phi(\textbf{p}_t)$, and $D(\cdot)$ is a predefined distance metric like the Euclidean distance. $L_{f\_att}$ ensures the gradual embedding of $\phi(\textbf{p}_t)$ closer to the hypersphere created with $\textbf{c}_t^k$ as center, facilitating feature adaptation.

To have a more discriminative patch descriptor network $\phi(\cdot)$, CFA incorporates hard negative features, which are the $K\!\!+\!\!j$-th nearest neighbors of $\phi(\textbf{p}_t)$, denoted as $\textbf{c}_t^j$. The contrastive supervision term, $L_{f\_rep}$, is introduced to repel $\textbf{p}_t$ from the hypersphere created with $\textbf{c}_t^j$ as the center, and is formulated as:
\begin{equation}
    L_{f\_rep} = \frac{1}{TJ}\sum_{t=1}^{T}\sum_{j=1}^{J}\max\Bigg(0,r^2-D\left(\phi(\textbf{p}_t),\textbf{c}_t^j\right)-\alpha\Bigg),
\end{equation}
where $J$ denotes the total number of hard negative features used for contrastive supervision, and $\alpha$ is a term used to balance the contribution of $L_{f\_rep}$ in the overall loss function, $L_{\operatorname{cfa}}$. 

$L_{\operatorname{cfa}}$ is the sum of these two loss terms, i.e.:
\begin{equation}
    L_{\operatorname{cfa}} = L_{f\_att} + L_{f\_rep}.
\end{equation}

Minimizing $L_{\operatorname{cfa}}$ optimizes the weights of the patch descriptor network $\phi(\cdot)$, ensuring densely clustered patch features and aiding in distinguishing normal and abnormal features.

Finally, 
as the minimum distance between $\phi(\textbf{p}_t)$ and memorized features in $C$, $D(\phi(\textbf{p}_t),\textbf{c}_t^1)$ quantifies the abnormality of $\textbf{p}_t$ and is used to make the anomaly score map. Afterward, the anomaly score map is properly interpolated to the same resolution as the input sample and smoothed using Gaussian smoothing as the post-processing step to provide the final anomaly score map.

\subsection{Regularized Discriminator}

\begin{figure}
\centering
\includegraphics[width=0.6\columnwidth]{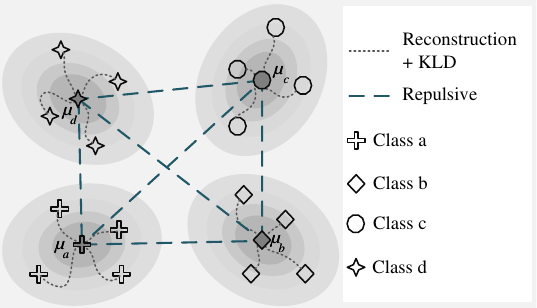}
\caption{Forces caused by reconstruction, KLD, and repulsive losses in RD-VAE.}
\label{rd-vae}
\end{figure}

The Regularized Discriminative Variational Auto-Encoder (RD-VAE) was introduced in \cite{passalis2020variance} in the context of content-based image retrieval. RD-VAE modifies the training process of VAEs for forcing similar samples to be grouped into distinct and well-separated clusters based on their classes using $N_c$ (equal to the number of the classes) individual Gaussian distributions in the representation space, each having a distinct mean $\bm{\mu}_{m}, \:m \in \{1,2,..., N_{c}\}$ and an identity covariance matrix. Figure \ref{rd-vae} illustrates a schematic 2D representation of RD-VAE within the latent space.
To do this, the loss function of RD-VAE $L_{\operatorname{rd\_vae}}$ is a combination of a reconstructive loss (e.g., $mse_{loss}$), a supervised Kullback–Leibler divergence loss $L_{KLD}$, and a distribution repulsive loss $L_{d\_rep}$.

Considering to have a collection of $N_s$ samples $X=\{\textbf{x}_{1},\textbf{x}_{2},...,\textbf{x}_{N_s}\}$, for each given sample $\textbf{x}_i$, the encoder is used to predict the mean $\mu_{Q}(\textbf{x}_i)$ and the covariance matrix $\Sigma_{Q}(\textbf{x}_i)$. Therefore, to have $N_c$ Gaussian distributions, the supervised $L_{KLD}$ loss is formulated as:
\begin{equation}
L_{KLD} = \sum_{\textbf{x}_i}\Bigl((\mu_{Q}(\textbf{x}_i) - \bm{\mu}_{l_i})^{T}(\mu_{Q}(\textbf{x}_i) - \bm{\mu}_{l_i}) + tr(\Sigma_{Q}(\textbf{x}_i)) - \log\det(\Sigma_{Q}(\textbf{x}_i)) - m\Bigr),\label{KLD_s}
\end{equation}
where $tr(\cdot)$ is the matrix trace operator, and $m$ and $l_i\in\{1,2,..,N_{c}\}$ are the dimensionality of the latent space and the class label of the sample $\textbf{x}_i$, respectively.

$L_{d\_rep}$ enforces a minimum distance $\rho$ between the means of different class distributions, and it is formulated as:
\begin{equation}
L_{d\_rep}\!=\!
\frac{1}{\rho}\sum_{\textbf{x}_i}\sum_{\textbf{x}_j \ne \textbf{x}_i}{max\left(0,\rho \!-\!\parallel \bm{\mu}_{l_i}\!-\!\bm{\mu}_{l_j} \parallel_{2}^{2}\right)\!^2}.\label{Rep}
\end{equation}

$L_{\operatorname{rd\_vae}}$ combines the above-described loss terms as follows:
\begin{equation}
L_{\operatorname{rd\_vae}}\!=\!mse_{loss}\!+\!\alpha_{kl}L_{KLD}\!+\!L_{d\_rep},\label{L_rd}
\end{equation}
where $\alpha_{kl}$ is a hyperparameter controlling the importance of the KL-divergence term in the optimization problem.

\section{Regularized Discriminative CFA (RD-CFA)}
CFA has demonstrated remarkable performance in anomaly detection tasks where normal samples come from one class across diverse datasets \cite{lee2022cfa}. However, its application to multi-class anomaly detection poses challenges due to its inability to generalize in such scenarios. The preceding section outlined that CFA relies on memorizing compressed features from all normal samples in a training set stored within a memory bank, and the patch descriptor is then trained to calculate features for normal inputs which are close to these memorized features. In a multi-class dataset, the memory bank needs to contain features from all classes, making effective training of the patch descriptor for all classes a complex task, as different classes may contain very different patterns that need to be represented adequately well for discriminating them from anomalous inputs in inference. To address this limitation, we introduce the discriminative capabilities of the RD-VAE method into the CFA model to enhance its performance in multi-class anomaly detection.

\begin{figure*}
\centering
\includegraphics[width=\textwidth]{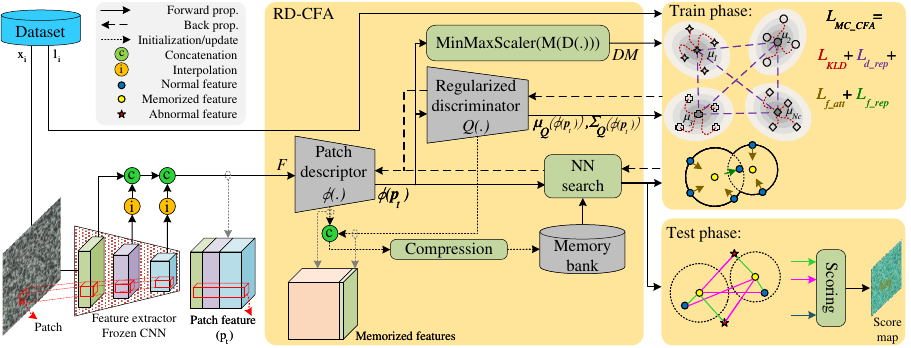}
\caption{Overall structure of the proposed RD-CFA model.}
\label{MCCFA}
\end{figure*}

To this end, as shown in Figure \ref{MCCFA}, the features $F\in\mathbb{R}^{D\times H\times W}$ extracted by the (frozen) CNN are introduced to the patch descriptor network $\phi(\cdot): \mathbb{R}^D \rightarrow \mathbb{R}^{D'}$ to generate target-oriented features $\phi(\textbf{p}_t)\in\mathbb{R}^{D'\times H\times W}$. Subsequently, these target-oriented features are processed by the regularized discriminator $Q(\cdot)$ to calculate the mean $\mu_{Q}(\phi(\textbf{p}_t))$ and the covariance matrix $\Sigma_{Q}(\phi(\textbf{p}_t))$ for each target-oriented patch feature. 

While in the RD-VAE method the $\bm{\mu}_{m}$s corresponding to all classes are enforced to be distant from each other with uniform enforcement, we argue that for multi-class anomaly detection, this discrimination should be proportionate to the dissimilarities between the classes. Therefore, before applying $L_{d\_rep}$ to the $\bm{\mu}_{l_i}$s, we need to calculate the dissimilarities between the classes and adjust repulsive enforcements proportionally. To achieve this, we feed the model with $N$ randomly selected samples (for instance, $N$ represents the batch size) from various classes in the dataset to evaluate the correlation among their features. Consequently, we obtain $N\times D'\times H\times W$ target-oriented patch features. By defining a pairwise distance function $D(\cdot, \cdot): \mathbb{R}^{ND'HW} \times \mathbb{R}^{ND'HW} \rightarrow \mathbb{R}^{NHW \times NHW}$ to compute distances between all patch features, employing a mean function $M(.): \mathbb{R}^{NHW\times NHW} \rightarrow \mathbb{R}^{N\times N}$ to average over all patches of each input sample, and utilizing the $MinMaxScaler$ normalization function as $MinMaxScaler(M_{ij})=(M_{ij}-M_{min})/(M_{max}-M_{min})$, we can create a dissimilarity matrix $DM$ as follows:

\begin{equation}
    DM=MinMaxScaler\left(M(D(\phi(\textbf{p}),\phi(\textbf{p})))\right).
\end{equation} 

The elements in this matrix quantify the level of correlation between different classes. Hence, it is employed in $L_{d\_rep}$ to weigh the repulsive enforcement based on the dissimilarities proportionally. Consequently, we use the following modified version of $L_{KLD}$ and $L_{d\_rep}$:
\begin{equation}
L_{KLD} = \sum_{\textbf{x}_i}\sum_{\textbf{p}_t}\Bigl((\mu_{Q}(\textbf{p}_t) - \bm{\mu}_{l_i})^{T}(\mu_{Q}(\textbf{p}_t) - \bm{\mu}_{l_i}) + tr(\Sigma_{Q}(\textbf{p}_t)) - \log\det(\Sigma_{Q}(\textbf{p}_t)) - m\Bigr),
\end{equation}
\begin{equation}
L_{d\_rep}\!=\!\frac{1}{\rho} 
\sum_{\textbf{x}_i}\sum_{\textbf{x}_j}\max\!\left(0,DM_{ij}\cdot(\rho\!-\!\|\bm{\mu}_{l_i}\!-\!\bm{\mu}_{l_j}\|_{2}^{2})\right)^{2}.
\end{equation}

The total loss function is defined as:
\begin{equation}
L_{\operatorname{rd\_cfa}} = \alpha_{kl} L_{KLD} + \alpha_{dr} L_{d\_rep} + L_{f\_att} + L_{f\_rep},
\end{equation}
where the coefficients $\alpha_{kl}$ and $\alpha_{dr}$ are hyperparameters used to adjust the influence of the corresponding terms in the total loss function. 

Although the proposed process leads to having proportionally discriminated features, these features' impact must be effectively transferred to the anomaly detection part of the model, i.e.,  the memory bank. To achieve this, we propose two modifications to the usage of the memory bank:
\begin{enumerate}
    \item Concatenating the $\mu_{Q}(\textbf{p}_t)$ and $\Sigma_{Q}(\textbf{p}_t)$ features 
    with the target-oriented feature of corresponding patch $\phi(\textbf{p}_t)$ 
    to be memorized in the memory bank.
    \item In addition to initializing the memory bank at the beginning of the training, we update it after each epoch.
\end{enumerate}

The first change provides additional features to the memory bank, enabling discrimination over features related to different classes and facilitating the training of the patch descriptor network $\phi(\cdot)$ in a multi-class dataset. As these added features in the memory bank better capture class discrimination during training, updating the memory bank results in memorizing features that more accurately represent class discrimination.

\section{Evaluation}
This section presents experiments conducted to assess the performance of the proposed RD-CFA method. The experiments were conducted on two widely used public datasets:
\begin{itemize}
\item The \textbf{MVTec AD} \cite{bergmann2019mvtec} dataset which contains 5,354 industrial samples distributed across 15 classes, including items like Bottles, Cables, Capsules, Carpets, Grids, Hazelnuts, Leather, Metal nuts, Pills, Screws, Tiles, Toothbrushes, Transistors, Wood, and Zippers.
\item The \textbf{BeanTech AD} \cite{mishra2021vt} dataset which contains 2,542 real-world industrial samples, encompassing both body and surface defects. The samples are categorized into three distinct classes.
\end{itemize}

\begin{figure}[!t]
\centering
\includegraphics[width=0.65\columnwidth]{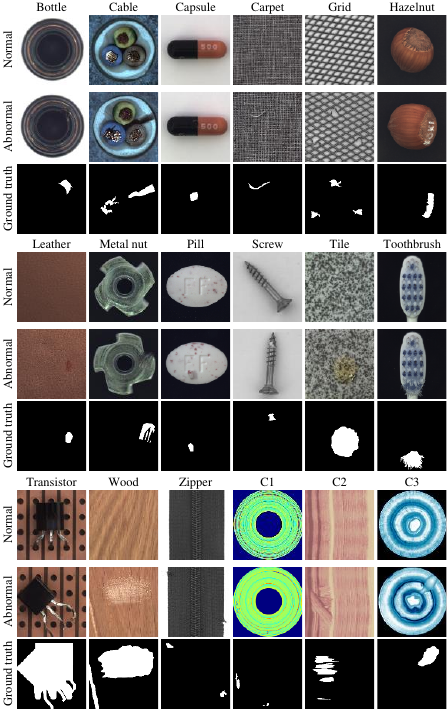}
\caption{Normal, abnormal, and ground truth samples from MVTec AD and BeanTech AD datasets.}
\label{mvtec}
\end{figure}

A normal and abnormal sample of each class in both MVTec AD and BeanTech AD datasets, along with their ground truth masks, are shown in Figure \ref{mvtec}. We combined all training and validation samples from different classes within each dataset to create the training and validation sets for the corresponding multi-class anomaly detection tasks. The trained models were evaluated separately on each test set corresponding to individual classes during the testing phase. 
To evaluate the performance of the proposed method, we employed the Area Under the Receiver Operator Curve (AUROC) as our evaluation metric, and we compared it with the performance of eight leading contemporary anomaly detection methods (DRÆM \cite{zavrtanik2021draem}, Patch Distribution Modeling (PaDiM) \cite{defard2021padim}, FastFlow \cite{yu2021fastflow}, CFA \cite{lee2022cfa}, CFLOW \cite{gudovskiy2022cflow},  EfficientAD \cite{batzner2023efficientad}, Deep Features Modeling (DFM) \cite{ahuja2019probabilistic}, and Reverse Distillation \cite{deng2022anomaly})  \footnote{The results for these methods were generated using the Anomalib library (\url{https://github.com/openvinotoolkit/anomalib}).}.

\subsection{Experimental Setup}
All CNNs used in the experiments were pre-trained on ImageNet \cite{deng2009imagenet}, and feature maps corresponding to $\{C2, C3, C4\}$ from intermediate layers as in \cite{lin2017feature} are extracted and used to acquire multiscale features. For RD-CFA, Wide-ResNet50 \cite{zagoruyko2016wide} is used as the backbone network as it showed the best performance in the original study \cite{lee2022cfa}. The hyperparameters associated with CFA were configured to match the values specified in \cite{lee2022cfa}. Using a grid search strategy with search choices of $\rho\in\{5,10,20,50\}$, $\alpha_{kl}\in\{0.1,0.25,0.5,0.75,1\}$, and $\alpha_{dr}\in\{0.1,0.25,0.5,0.75,1\}$, these hyperparameters are set to 10, 0.5, and 0.1, respectively. All the experiments are conducted five times, and the reported results correspond to the average performance values.

\subsection{Quantitative Results}
The quantitative evaluation of the competing methods is conducted on both datasets. Additionally, for the MVTec AD dataset, results from the UniAD method are included directly from \cite{you2022unified}, as they are only available on this dataset.

Tables \ref{res_I_mv} and \ref{res_P_mv} provide the performance of the competing methods on anomaly detection and localization across all fifteen classes in the MVTec AD dataset. These tables provide the average performance for both object and texture classes and the overall average performance for each method. The proposed RD-CFA method substantially enhances multi-class anomaly detection compared to the original CFA method. While RD-CFA does not achieve the best results for every individual class, it outperforms the competing methods in terms of average performance across object classes and the overall average for anomaly detection. Furthermore, it also outperforms other methods across all averages for anomaly localization.

\begin{table*}[!htbp]
\setlength{\abovecaptionskip}{0pt} 
\setlength{\belowcaptionskip}{0pt} 
\caption{Anomaly detection results with AUROC (\%) metric on MVTec AD  dataset (including each class and averages over object classes, texture classes, and the overall average).}\label{res_I_mv}
\centering
\footnotesize
\begin{tabular}{p{10mm}p{7mm}p{6mm}p{8.5mm}p{4.5mm}p{8mm}p{8mm}p{4.5mm}p{11.5mm}p{7mm}p{9mm}}
\hline
\rowcolor[HTML]{808080} 
Category     & DRÆM & PaDiM & FastFlow & CFA  & CFLOW & Efficient AD & DFM  & Reverse   Distillation & UniAD & RD-CFA (Ours) \\ \hline
bottle       & 78.8 & 98.0  & 98.2     & 87.9 & 99.9  & 94.6        & \textbf{99.7} & 65.5                   & \textbf{99.7} & 98.3          \\ \hline
\rowcolor[HTML]{F2F2F2} 
cable        & 59.8 & 71.2  & 80.0     & 76.0 & 41.0  & 78.8        & 76.0          & 87.1                   & 95.2          & \textbf{95.9} \\ \hline
capsule      & 51.5 & 72.3  & 82.0     & 75.9 & 81.4  & 55.4        & 92.3          & 91.1                   & 86.9          & 92.6          \\ \hline
\rowcolor[HTML]{F2F2F2} 
carpet       & 94.3 & 74.1  & 78.6     & 84.1 & 83.5  & 91.3        & 86.8          & 98.3                   & \textbf{99.8} & 98.2          \\ \hline
grid         & 54.9 & 71.5  & 84.9     & 78.3 & 72.3  & 97.7        & 51.5          & 97.8                   & \textbf{98.2} & 97.4          \\ \hline
\rowcolor[HTML]{F2F2F2} 
hazelnut     & 79.3 & 97.5  & 75.6     & 81.4 & 94.7  & 84.6        & 77.8          & \textbf{100}           & 99.8          & 97.5          \\ \hline
leather      & 84.1 & 95.7  & 93.9     & 92.1 & 98.2  & 83.4        & 85.1          & 99.9                   & \textbf{100}  & \textbf{100}  \\ \hline
\rowcolor[HTML]{F2F2F2} 
metal nut    & 69.6 & 80.4  & 86.3     & 88.6 & 91.1  & 73.1        & 89.4          & 99.0                   & \textbf{99.2} & \textbf{99.2} \\ \hline
pill         & 59.3 & 59.9  & 76.5     & 72.9 & 32.6  & 86.2        & 51.9          & 95.3                   & 93.7          & \textbf{95.6} \\ \hline
\rowcolor[HTML]{F2F2F2} 
screw        & 79.3 & 50.9  & 55.7     & 77.0 & 39.5  & 52.3        & 73.8          & 93.2                   & 87.5          & \textbf{93.5} \\ \hline
tile         & 86.6 & 85.0  & 88.2     & 90.4 & 95.1  & 95.7        & 83.4          & \textbf{99.4}          & 99.3          & 98.2          \\ \hline
\rowcolor[HTML]{F2F2F2} 
toothbrush   & 67.2 & 86.1  & 64.1     & 73.8 & 53.3  & 67.5        & \textbf{96.9} & 95.0                   & 94.2          & 93.2          \\ \hline
transistor   & 59.3 & 81.0  & 83.1     & 78.5 & 53.0  & 74.4        & 57.1          & 93.0                   & \textbf{99.8} & 98.0          \\ \hline
\rowcolor[HTML]{F2F2F2} 
wood         & 87.9 & 95.8  & 97.0     & 93.0 & 96.1  & 88.6        & 86.3          & 99.3                   & 98.6          & \textbf{99.6} \\ \hline
zipper       & 72.3 & 72.7  & 87.2     & 86.1 & 88.2  & 91.0        & 96.2          & 96.4                   & 95.8          & \textbf{97.0} \\ \hline
\rowcolor[HTML]{BFBFBF} 
avg. obj.  & 67.6 & 77.0  & 78.9     & 79.8 & 67.5  & 75.8        & 81.1          & 91.6                   & 95.2          & \textbf{96.1} \\ \hline
\rowcolor[HTML]{BFBFBF} 
avg. tex. & 80.9 & 81.6  & 87.2     & 86.5 & 86.8  & 93.3        & 77.0          & 98.7                   & \textbf{99.0} & 98.7          \\ \hline
\rowcolor[HTML]{BFBFBF} 
avg. total   & 72.3 & 79.5  & 82.1     & 82.4 & 74.7  & 81.0        & 80.3          & 94.0                   & 96.5          & \textbf{96.9} \\ \hline
\end{tabular}

\end{table*}

\begin{table*}[!htbp]
\setlength{\abovecaptionskip}{0pt} 
\setlength{\belowcaptionskip}{0pt} 
\caption{Anomaly localization results with AUROC (\%) metric on MVTec AD  dataset (including each class and averages over object classes, texture classes, and the overall average).}\label{res_P_mv}
\centering
\footnotesize
\begin{tabular}{p{10mm}p{7mm}p{6mm}p{8.5mm}p{4.5mm}p{8mm}p{8mm}p{4.5mm}p{11.5mm}p{7mm}p{9mm}}
\hline
\rowcolor[HTML]{808080} 
Category     & DRÆM & PaDiM & FastFlow & CFA  & CFLOW & Efficient AD & DFM  & Reverse   Distillation & UniAD & RD-CFA (Ours) \\ \hline
bottle       & 77.5 & 96.4  & 92.8     & 89.8 & 97.4          & 90.0        & 96.9          & 95.9                   & 98.1          & \textbf{98.4} \\ \hline
\rowcolor[HTML]{F2F2F2} 
cable        & 42.8 & 85.6  & 90.7     & 86.8 & 79.9          & 75.2        & \textbf{97.7} & 78.9                   & 97.3          & 96.9          \\ \hline
capsule      & 72.8 & 97.3  & 95.2     & 91.2 & 97.8          & 91.0        & 97.9          & \textbf{98.6}          & 98.5          & \textbf{98.6} \\ \hline
\rowcolor[HTML]{F2F2F2} 
carpet       & 61.8 & 92.3  & 93.3     & 93.0 & \textbf{98.6} & 95.7        & 97.5          & 97.8                   & 98.5          & 97.7          \\ \hline
grid         & 46.9 & 60.4  & 88.3     & 88.5 & 93.4          & 90.5        & 90.7          & 95.0                   & \textbf{96.5} & 94.3          \\ \hline
\rowcolor[HTML]{F2F2F2} 
hazelnut     & 76.6 & 96.9  & 94.1     & 86.6 & 97.2          & 93.5        & 98.1          & 98.7                   & 98.1          & \textbf{98.9} \\ \hline
leather      & 56.2 & 97.4  & 98.2     & 93.3 & \textbf{99.0} & 94.3        & 98.2          & \textbf{99.0}          & 98.8          & \textbf{99.0} \\ \hline
\rowcolor[HTML]{F2F2F2} 
metal nut    & 76.7 & 87.8  & 95.3     & 90.5 & 93.3          & 93.9        & 96.3          & 94.5                   & 94.8          & \textbf{96.8} \\ \hline
pill         & 78.0 & 90.1  & 91.0     & 88.5 & 93.4          & 96.0        & 96.4          & 97.2                   & 95.0          & \textbf{98.4} \\ \hline
\rowcolor[HTML]{F2F2F2} 
screw        & 84.3 & 94.4  & 87.0     & 87.2 & 95.3          & 87.9        & 97.8          & \textbf{99.3}          & 98.3          & 97.1          \\ \hline
tile         & 74.8 & 77.6  & 91.6     & 89.6 & \textbf{96.2} & 84.7        & 90.1          & 93.4                   & 91.8          & 94.7          \\ \hline
\rowcolor[HTML]{F2F2F2} 
toothbrush   & 84.6 & 97.1  & 93.4     & 93.7 & 96.7          & 92.3        & 98.4          & 98.6                   & 98.4          & \textbf{98.7} \\ \hline
transistor   & 52.0 & 90.6  & 89.5     & 86.9 & 79.9          & 84.6        & \textbf{98.5} & 83.1                   & 97.9          & 96.5          \\ \hline
\rowcolor[HTML]{F2F2F2} 
wood         & 67.7 & 89.0  & 91.6     & 80.6 & 93.5          & 79.3        & 89.3          & 95.0                   & 93.2          & \textbf{95.1} \\ \hline
zipper       & 66.3 & 92.2  & 93.6     & 92.2 & 95.1          & 93.1        & 95.9          & \textbf{98.5}          & 96.8          & 94.9          \\ \hline
\rowcolor[HTML]{BFBFBF} 
avg. obj.  & 71.2 & 92.9  & 92.3     & 89.3 & 92.6          & 75.8        & 97.4          & 94.4                   & 97.3          & \textbf{97.5} \\ \hline
\rowcolor[HTML]{BFBFBF} 
avg. tex. & 62.8 & 79.9  & 91.2     & 87.9 & 95.4          & 93.3        & 91.9          & 95.3                   & 95.0          & \textbf{95.5} \\ \hline
\rowcolor[HTML]{BFBFBF} 
avg. total   & 67.9 & 89.7  & 92.4     & 89.2 & 93.8          & 89.5        & 96.0          & 94.9                   & 96.8          & \textbf{97.1} \\ \hline
\end{tabular}
\end{table*} 
\begin{table*}
\caption{Anomaly detection results with AUROC (\%) metric on BeanTech dataset (including each class, and the overall average).}\label{res_I_be}
\centering
\footnotesize
\begin{tabular}
{p{10mm}p{7mm}p{6mm}p{9mm}p{5mm}p{8mm}p{8mm}p{5mm}p{12mm}p{9mm}}
\hline
\rowcolor[HTML]{808080} 
Category     & DRÆM & PaDiM & FastFlow & CFA  & CFLOW & Efficient AD & DFM  & Reverse   Distillation & RD-CFA (Ours) \\ \hline
Class 1   & 97.2 & \textbf{99.0} & 93.7     & 69.7 & 98.1  & 91.7        & 97.0         & 76.9                   & 94.3          \\ \hline
\rowcolor[HTML]{F2F2F2} 
Class   2 & 88.1 & 80.9          & 80.2     & 74.6 & 84.1  & 83.7        & 80.3         & 87.2                   & \textbf{88.6} \\ \hline
Class 3   & 75.9 & 98.9          & 95.4     & 70.0 & 99.7  & 86.0        & \textbf{100} & \textbf{100}           & 98.8 \\ \hline
\rowcolor[HTML]{BFBFBF} 
Average   & 87.1 & 93.0          & 89.8     & 71.4 & 93.6  & 87.1        & 92.4         & 88.0                   & \textbf{93.9} \\ \hline
\end{tabular}
\end{table*}

\begin{table*}
\caption{Localization detection results with AUROC (\%) metric on BeanTech dataset (including each class, and the overall average).}\label{res_P_be}
\centering
\footnotesize
\begin{tabular}{p{10mm}p{7mm}p{6mm}p{9mm}p{5mm}p{8mm}p{8mm}p{5mm}p{12mm}p{9mm}}
\hline
\rowcolor[HTML]{808080} 
Category     & DRÆM & PaDiM & FastFlow & CFA  & CFLOW & Efficient AD & DFM  & Reverse   Distillation & RD-CFA (Ours) \\ \hline
Class 1   & 65.3 & 96.6  & 91.4     & 92.0 & 94.0  & 90.3        & 97.4 & 96.9                   & \textbf{98.9} \\ \hline
\rowcolor[HTML]{F2F2F2} 
Class   2 & 78.2 & 94.9  & 95.5     & 92.1 & 95.3  & 92.0        & 94.7 & 96.7                   & \textbf{97.9} \\ \hline
Class 3   & 85.9 & 99.3  & 99.0     & 92.7 & 99.3  & 94.5        & 99.6 & \textbf{99.7}          & 99.1          \\ \hline
\rowcolor[HTML]{BFBFBF} 
Average   & 76.5 & 96.9  & 95.3     & 92.3 & 96.2  & 92.3        & 97.2 & 97.8                   & \textbf{98.6} \\ \hline
\end{tabular}
\end{table*}

Tables \ref{res_I_be} and \ref{res_P_be} provide the performance of the competing methods across all three classes in the BeanTech AD dataset and the overall average performance. Once again, the results show substantial performance enhancement compared to the CFA method and a higher on average performance for the proposed RD-CFA method when compared to the competing methods.

\subsection{Ablation Studies}
To evaluate the effect of combining the features calculated by the regularized discriminator with those from the patch descriptor to be stored in the memory bank, and the influence of updating the memory bank throughout training, we conducted a set of experiments on both datasets, as shown in Table \ref{abl}. These results demonstrate the positive impact of both method enhancements on anomaly detection and localization for both datasets. The best performances are obtained when both strategies are implemented simultaneously.

\begin{table*}
\caption{Average detection and localization results with AUROC (\%) metric of the proposed method, according to added features and memory update policies on MVTec AD and BeanTch AD datasets.}\label{abl}
\centering
\footnotesize
\begin{tabular}{cccccc}
\hline
\rowcolor[HTML]{808080} 
\cellcolor[HTML]{808080}   &   \cellcolor[HTML]{808080}   & \multicolumn{2}{c}{\cellcolor[HTML]{808080}MVTec} & \multicolumn{2}{c}{\cellcolor[HTML]{808080}BeanTech} \\ 
\rowcolor[HTML]{808080} 
\multirow{-2}{*}{\cellcolor[HTML]{808080}\begin{tabular}[c]{@{}l@{}}Extra feature\end{tabular}} & \multirow{-2}{*}{\cellcolor[HTML]{808080}\begin{tabular}[c]{@{}c@{}}Update \\ memory bank\end{tabular}} & Detection  & Localization  & Detection  & Localization \\  \hline
     &    & 82.5  & 85.5  & 86.1  & 93.3 \\ \hline
\rowcolor[HTML]{F2F2F2} 
\checkmark    &    & 86.7  & 92.1  & 87.7  & 96.3 \\ \hline
     & \checkmark    & 96.4  & 95.3  & 90.9  & 97.4 \\ \hline
\rowcolor[HTML]{F2F2F2} 
\checkmark    & \checkmark    & \textbf{96.9} & \textbf{97.1} & \textbf{93.9} & \textbf{98.6}  \\ \hline
\end{tabular}
\end{table*}

Furthermore, to evaluate the influence of aligning the $L_{d\_rep}$ loss with the dissimilarities among various classes, we conducted experiments on both datasets, as shown in Table \ref{abl2}. As can be seen, incorporating the dissimilarity matrix into the $L_{d\_rep}$ loss marginally enhances the overall performance.

\begin{table*}
\caption{Average detection and localization results with AUROC (\%) metric of the proposed method, according to added dissimilarity matrix policies on MVTec AD and BeanTch AD datasets.}\label{abl2}
\centering
\footnotesize
\begin{tabular}{ccccc}
\hline
\rowcolor[HTML]{808080} 
\cellcolor[HTML]{808080}    & \multicolumn{2}{c}{\cellcolor[HTML]{808080}MVTec} & \multicolumn{2}{c}{\cellcolor[HTML]{808080}BeanTech} \\  
\rowcolor[HTML]{808080} 
\multirow{-2}{*}{\cellcolor[HTML]{808080}\begin{tabular}[c]{@{}c@{}}Dissimilarity matrix\end{tabular}} & Detection  & Localization  & Detection  & Localization \\ \hline
    & \textbf{96.9}  & 96.8  & 93.5  & 98.1 \\ \hline
\rowcolor[HTML]{F2F2F2} 
\checkmark     & \textbf{96.9} & \textbf{97.1} & \textbf{93.9} & \textbf{98.6}  \\ \hline
\end{tabular}
\end{table*}

\subsection{Qualitative Results}
\begin{figure*}
\centering
\includegraphics[width=\textwidth]{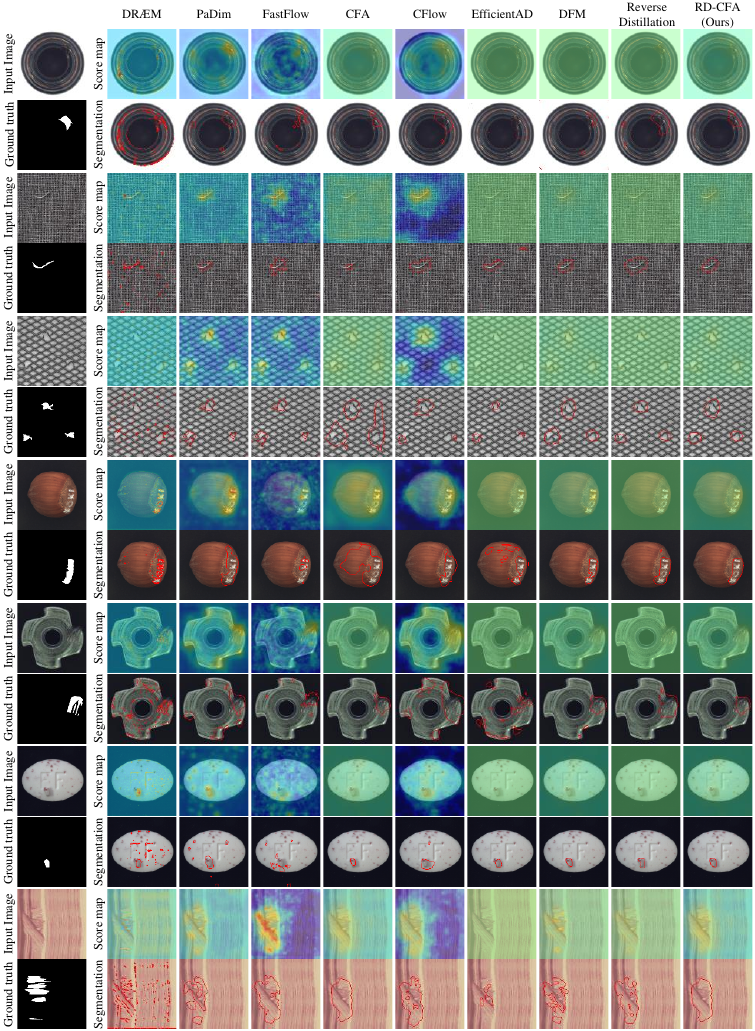}
\caption{Visualization of results of anomaly localization for random classes in MVTec AD and BeanTech datasets.}
\label{qual}
\end{figure*}

Figure \ref{qual} provides qualitative results of anomaly localization. These results feature randomly selected samples from various classes within the MVTec AD and BeanTech AD datasets obtained through our proposed method and eight other state-of-the-art techniques. The figure exhibits the input image, the ground truth, predicted score maps from each method, and the corresponding segmented abnormal areas.

As was argued, accurate discrimination among different classes is crucial in a multi-class anomaly detection scenario. Methods that lack this discrimination tend to produce false positive predictions, indicating an inability to focus on the specific normal features of each class. The results demonstrate that our proposed method excels in this aspect, producing segmented areas that primarily highlight defective parts with minimal false positives. This emphasizes the method's capability to effectively distinguish among various classes, providing a more precise assessment of abnormality levels within patches. Therefore, the consistent and accurate localization of abnormal areas across all products, even in challenging cases, underscores the high qualitative performance of our proposed method when compared to other techniques.

\section{Conclusions}
In this paper, we introduced RD-CFA for multi-class anomaly detection that incorporates the discriminative capabilities of Regularized Discriminative Variational Auto-Encoder (RD-VAE) to Coupled-hypersphere-based Feature Adaptation (CFA) to enable it to perform multi-class anomaly detection. We evaluated its performance on the widely used publicly available MVTec AD and BeanTech AD datasets and compared it to that of eight leading contemporary anomaly detection methods in both anomaly detection and localization tasks, showing consistent performance improvements.

\section*{Acknowledgments}

The research leading to the results of this paper received funding from the Innovation Fund Denmark as part of MADE FAST.

\section*{Statement}
\noindent The paper is under consideration at Pattern Recognition Letters.








\bibliography{ref}

\end{document}